\documentclass{article}
\usepackage{ijcai21}

\usepackage{times}
\usepackage{soul}
\usepackage{url}
\usepackage[hidelinks]{hyperref}
\usepackage[utf8]{inputenc}
\usepackage[small]{caption}
\usepackage{graphicx}
\usepackage{amsmath}
\usepackage{amsthm}
\usepackage{booktabs}
\usepackage{algorithm}
\usepackage{algorithmic}
\usepackage{bm}
\usepackage{mathtools}
\usepackage{multirow}
\usepackage{subfigure}
\usepackage{xcolor}
\usepackage{amssymb}
\usepackage{dblfloatfix}

\newcommand{\ie}{{\it i.e.}}

\newcommand{\wrt}{w.r.t. }

\renewcommand{\vec}[1]{\ensuremath{\mathbf{#1}}}

\title{Node-wise Localization of Graph Neural Networks}

\author{
Zemin Liu$^1$\and
Yuan Fang$^1$\and
Chenghao Liu$^{2}$\And
Steven C.H. Hoi$^{1,2}$\\
\affiliations
$^1$Singapore Management University, Singapore\\
$^2$Salesforce Research Asia, Singapore\\
\emails
\{zmliu, yfang\}@smu.edu.sg,
\{chenghao.liu, shoi\}@salesforce.com
}

\begin{document}

\maketitle

\begin{abstract}
Graph neural networks (GNNs) emerge as a powerful family of representation learning models on graphs.
To derive node representations, they utilize a global model that recursively aggregates information from the neighboring nodes.
However, different nodes reside at different parts of the graph in different local contexts, making their distributions vary across the graph. 
Ideally, how a node receives its neighborhood information should be a function of its local context, 
to diverge from the global GNN model shared by all nodes. 
To utilize node locality without overfitting, we propose a node-wise localization of GNNs by accounting for both global and local aspects of the graph.  
Globally, all nodes on the graph depend on an underlying global GNN to encode the general patterns across the graph;
locally, each node is localized into a unique model as a function of the global model and its local context. 
Finally, we conduct extensive experiments on four benchmark graphs, and consistently obtain promising performance surpassing the state-of-the-art GNNs.
\end{abstract}

\section{Introduction}

\begin{figure*}[t]
\centering
\includegraphics[width=0.9\linewidth]{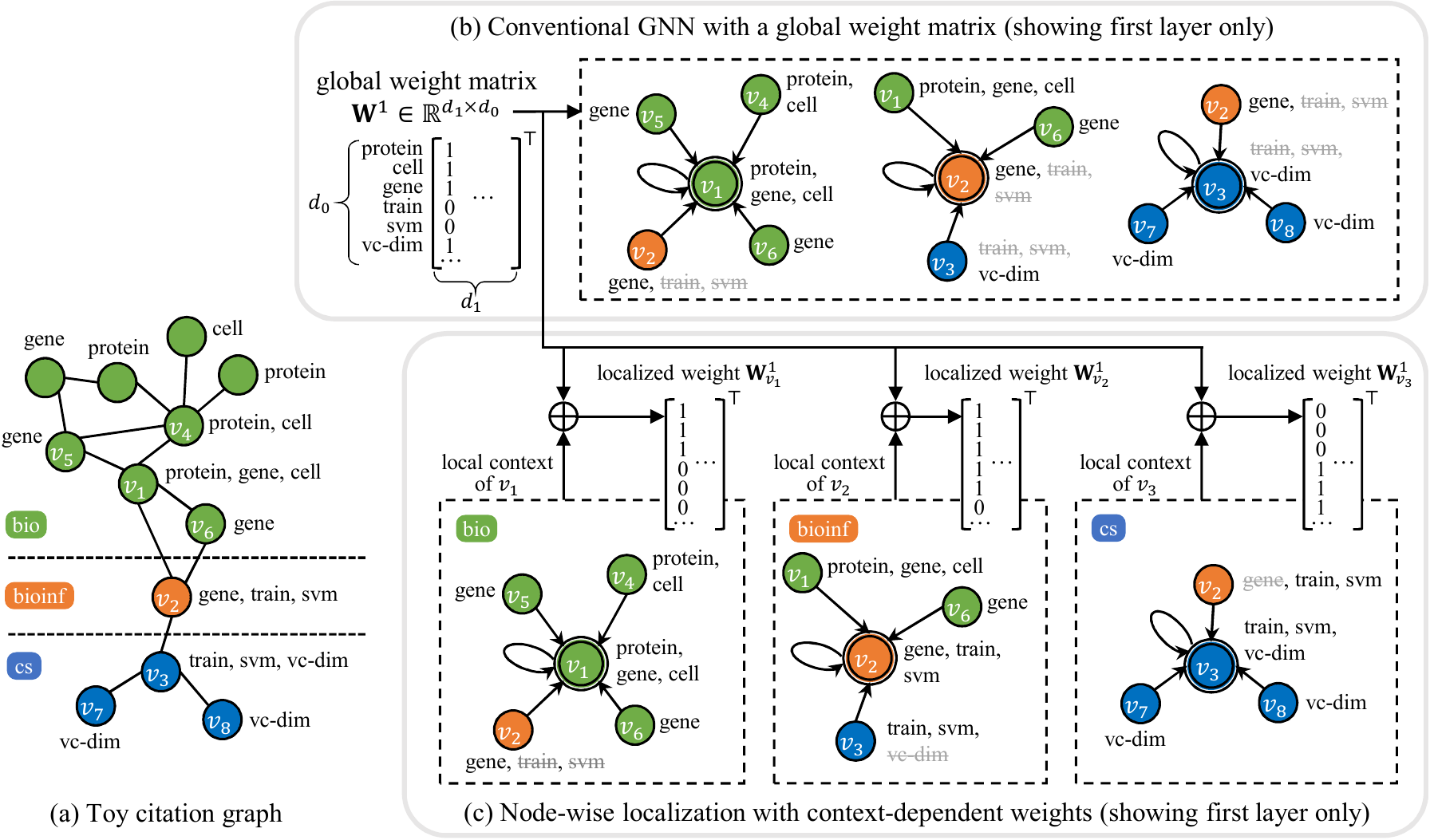}
\caption{Comparison of conventional GNNs and the proposed localization (best viewed in color).}
\label{fig.intro}
\end{figure*}

Graphs are powerful data structures to model various entities (\ie, nodes) and their interactions (\ie, edges) simultaneously. 
To learn the representations of nodes on a graph, graph neural networks (GNNs) \cite{wu2020comprehensive} have been proposed as a promising solution. 
Generally, state-of-the-art GNNs \cite{kipf2016semi,hamilton2017inductive,velivckovic2017graph} 
learn the representation of each node by recursively transferring and aggregating information from its receptive field, which is often defined as its set of neighboring nodes.
Consider a toy citation graph in Fig.~\ref{fig.intro}(a), consisting of papers in three areas: biology (\emph{bio}), bioinformatics (\emph{bioinf}) and computer science (\emph{cs}). To derive the representation of a node, say $v_1$, GNNs aggregate keyword features from not only $v_1$ itself, but also its neighbors, namely $v_2,v_4,v_5$ and $v_6$, as illustrated in Fig.~\ref{fig.intro}(b). Such neighborhood aggregation can be performed recursively for the neighbors as well in more layers, to fully exploit the graph structures. To extract useful representations from the neighbors, GNNs aim to learn the model parameters formulated as a sequence of weight matrices $\vec{W}^{1},\vec{W}^{2},\dots,\vec{W}^{l}$ in each layer.

However, the implicit assumption of a global weight matrix (in each layer) for the entire graph is often too strict. Nodes do not distribute uniformly over the graph, and are associated with different local contexts.
For instance, in Fig.~\ref{fig.intro}(c), node $v_1$ is associated with a \emph{bio} context, $v_2$ with a \emph{bioinf} context and $v_3$ with a \emph{cs} context. Different local contexts are characterized by different keyword features, such as ``gene'' and ``cell'' in \emph{bio}, ``gene'' and ``svm'' in \emph{bioinf}, as well as ``svm'' and ``vc-dim'' in \emph{cs}. Thus, 
a graph-level global weight matrix is inherently inadequate to express the varying importance of features at different localities. More specifically, a global weight matrix can be overly diffuse, for different nodes often have distinct optimal weight matrices, 
which tend to pull the model in many opposing directions. This may result in a biased model that centers its mass around the most frequent patterns while leaving others not well covered.
A natural question follows: \emph{Can we allow each node to be parameterized by its own weight matrix?} Unfortunately, 
this is likely to cause severe overfitting to local noises and suffer from significantly higher training cost.

In this work, to adapt to the local context of each node without overfitting, we localize the model for each node from a shared global model. As illustrated in Fig.~\ref{fig.intro}(c), a localized weight matrix $\vec{W}^{l}_{v_i}$ for each node $v_i$ can be derived from both the global weight $\vec{W}^{l}$ and the local context of $v_i$.
That is, $\vec{W}^{l}_{v_i}$ is a function of the global information and the local context.
Globally, all nodes depend on a common underlying model to encode the general patterns at the graph level. Locally, each node leverages its local context to personalize the common model into its unique localized weight at the node level. It is also useful to incorporate a finer-grained localization at the edge level, where information received through each edge of the target node can be further adjusted.
The proposed localization, which we call \emph{node-wise localized GNN} (LGNN), aims to strike a balance between the global and local aspects of the graph. 
Moreover, LGNN is agnostic of the underlying global model, meaning that it 
is able to localize any GNN that follows the paradigm of recursive neighborhood aggregation, and subsumes various conventional GNNs  
as its  limiting cases. 

In summary, our main contributions are three-fold. 
(1) We identify the need to adapt to the local context of each node in GNNs.
(2) We propose a node-wise localization of GNNs to capture both global and local information on the graph, and further discuss its connection to other works.
(3) We conduct extensive experiments on four benchmark datasets and show that LGNN  consistently outperforms prior art.

\section{Proposed Approach}

In this section, we introduce the proposed  approach LGNN, and discuss its connection to other works.

\subsection{General Formulation of Localization}

We start with an abstract definition of localizing a global model to fit local contexts.
Specifically, the localized model of an instance $v$ is a function of both the global model and the local context.
Let $\Theta$ denote the global model, and $C_v$ the local context of $v$.
Then the localized model for $v$ is
\begin{align}\label{eq:localization}
    \Theta_v = f(\Theta, C_v),
\end{align}
where $f$ can be any function such as a neural network. 
While we focus on graph data where $v$ is a node, the general formulation is also pertinent to other kinds of data 
when there are varying, non-uniform contexts associated with instances. In particular,
on a graph $G=(V,E)$ with a set of nodes $V$ and edges $E$, the local context of a node $v$ can be materialized as the neighbors of $v$, in addition to $v$ itself. That is,
\begin{align}
    C_v=\{v\} \cup \{u\in V:\langle v,u\rangle\in E\}.
\end{align}

\subsection{Localization of GNNs}

A typical GNN consists of multiple layers of recursive neighborhood aggregation. In the $l$-th layer, each node $v$ receives and aggregates information from its neighbors to derive its hidden representation $\vec{h}_v^{l} \in \mathbb{R}^{d_l}$,
with $d_l$ being the dimension of the $l$-the layer representation. Note that the initial representation $\vec{h}_v^{0}$ is simply the input feature vector of $v$. The aggregation is performed on the local context of $v$ which also covers the neighbors of $v$, as follows.
\begin{align}\label{eq:gnn}
\vec{h}^{l}_v= \sigma \left(\textsc{Aggr}\left(\left\{\vec{W}^{l}\vec{h}^{l-1}_u : \forall u \in C_v \right\}\right)  \right),
\end{align}
where $\vec{W}^{l} \in \mathbb{R}^{d_l \times d_{l-1}}$ is a weight matrix in the $l$-th layer, $\sigma(\cdot)$ is an activation function, and \textsc{Aggr}$(\cdot)$ is an aggregation function. Different GNNs differ in the choice of the aggregation function. For instance, GCN \cite{kipf2016semi} uses an aggregator roughly equivalent to mean pooling \cite{hamilton2017inductive}, GAT \cite{velivckovic2017graph} uses an attention-weighted mean aggregator, and GIN \cite{xu2018powerful} uses a multi-layer perceptron (MLP).

\paragraph{Node-level localization.}
In the above setup, we have a global model $\vec{W}^{l}$ for all nodes in the $l$-th layer. 
At the node level, the global model is localized to adapt to the local context of each node. Specifically, we modulate the global weight $\vec{W}^{l}$ by a node-specific transformation through scaling and shifting. The localized weight matrix of node $v$ is given by
\begin{align} \label{node_level_localization}
\vec{W}^{l}_v = \vec{W}^{l} \odot \left[\left(\vec{a}^{l}_v\right)_{\times d_{l}}\right]^\top 
+ \left[\left(\vec{b}^{l}_v\right)_{\times d_{l}}\right]^\top,    
\end{align}
where $\vec{a}^{l}_v,\vec{b}^{l}_v \in \mathbb{R}^{d_{l-1}}$ are $v$-specific vectors for scaling and shifting, respectively.
Here the notation $[(\vec{x})_{\times n}]$ represents a matrix of $n$ columns all of which are identical to the vector $\vec{x}$, and
$\odot$ denotes element-wise multiplication. We essentially transform each row of the global  matrix $\vec{W}^{l}$
by $\vec{a}^{l}_v$ and $\vec{b}^{l}_v$ in an element-wise manner, to generate the localized weight $\vec{W}^{l}_v$, 
so that various importance levels are assigned to each feature dimension of the node embedding $\vec{h}^{l-1}_u$.

It is important to recognize that the node-specific transformation should not be directly learned, for two reasons. First, directly learning them  significantly increases the number of model parameters especially in the presence of a large number of nodes, causing overfitting. Second, node-wise localization is a function of both the global model and the local context as formulated in Eq.~\eqref{eq:localization}, in order to better capture the local information surrounding each node.  
Thus, for node $v$, we propose to generate the $v$-specific transformation in the $l$-th layer from its local contextual information $\vec{c}^{l}_v$. A straightforward recipe for $\vec{c}^{l}_v$ is to pool the $(l-1)$-th layer representations of the nodes in the local context $C_v$:
\begin{align} \label{film_node_neighbors}
    \vec{c}^{l}_v=\textsc{Mean}\left(\left\{\vec{h}^{l-1}_u:\forall u \in C_v\right\}\right),
\end{align}
in which we adopt the mean pooling although other forms of pooling can  be substituted for it. Given the local contextual information $\vec{c}^{l}_v \in \mathbb{R}^{d_{l-1}}$, we further utilize a dense layer, shared by all nodes, to generate the $v$-specific transformation as follows.
\begin{align} \label{film_node}
\vec{a}^{l}_{v} = \sigma\left(\vec{M}^{l}_{a} \vec{c}^{l}_v\right) + \vec{1},\quad
\vec{b}^{l}_{v} = \sigma\left(\vec{M}^{l}_{b} \vec{c}^{l}_v\right),
\end{align}
where $\vec{M}^{l}_{a},\vec{M}^{l}_{b}\in \mathbb{R}^{d_{l-1}\times d_{l-1}}$ are learnable parameters shared by all nodes, and $\vec{1}$ is a vector of ones to ensure that the scaling factors are centered around one.
Note that if the dimension $d_{l-1}$ is too large, such as that of the high-dimensional raw features handled in the first layer of a GNN, we could employ two dense layers 
and set the first one with a smaller number of neurons. 

\paragraph{Edge-level localization.}
The node-level localization of the global weight matrix is coarse-grained, as the same localized weight is applied on all neighbors of the target node. 
That is, the target node receives information through each of its edges uniformly. 
To enable a finer-grained localization, we further modulate how information propagates at the edge level. 

Consider a node $v$ with edges $\{\langle u,v\rangle : \forall u \in C_v\}$. In the $l$-th layer of the GNN, let $\vec{c}^{l}_{ u,v }$ denote the local contextual information of the edge $\langle u,v\rangle$, which is given by its two ends. Specifically, we concatenate the $(l-1)$-th layer representations of nodes $u$ and $v$, as follows.
\begin{align} \label{film_edge_neighbors}
\vec{c}^{l}_{  u,v  } =  \textsc{Concat}\left(\vec{h}^{l-1}_v, \vec{h}^{l-1}_u\right).
\end{align}
To implement the edge-level localization, we similarly adopt an edge-specific transformation through scaling and shifting.
Specifically, to derive the $l$-th layer representation of $v$, we scale and shift the information from each edge $\langle u,v\rangle$ during aggregation, by rewriting Eq.~\eqref{eq:gnn} as follows.
\begin{align} \label{eq:lgnn}
\hspace{-1mm}\vec{h}^{l}_v = \sigma\hspace{-.5mm}\left(\textsc{Aggr}\hspace{-.5mm}\left(\hspace{-.5mm}\left\{\vec{W}^{l}_v\vec{h}^{l-1}_u\odot \vec{a}^{l}_{  u,v } + \vec{b}^{l}_{  u,v }: \forall u \in C_v\right\}\hspace{-.5mm}\right)\hspace{-.5mm}\right),\hspace{-1mm}
\end{align}
where $\vec{a}^{l}_{ u,v }, \vec{b}^{l}_{  u,v } \in \mathbb{R}^{d_l}$ are edge $\langle u,v\rangle$-specific vectors for scaling and shifting, respectively.
Like the node-specific transformation, we generate the edge-specific transformation with a dense layer  given by
\begin{align} \label{film_edge}
\vec{a}^{l}_{  u,v  } = \sigma\left(\vec{N}^{l}_{a} \vec{c}^{l}_{  u,v }\right) + \vec{1}, \quad
\vec{b}^{l}_{  u,v  } = \sigma\left(\vec{N}^{l}_{b} \vec{c}^{l}_{  u,v }\right),
\end{align}
where $\vec{N}^{l}_{a},\vec{N}^{l}_{b}\in \mathbb{R}^{d_{l}\times 2d_{l-1}}$ are learnable parameters shared by all edges.

\subsection{Semi-supervised Node Classification}
In the benchmark task of semi-supervised node classification,
each node belongs to one of the pre-defined classes $\{1,2,\ldots,K\}$. However, we only know the class labels of a subset of nodes $V_Y \subset V$, called the labeled nodes. The goal is to predict the class labels of the remaining nodes. Following standard practice \cite{kipf2016semi}, for $K$-way classification we set the dimension of the output layer to $K$, and apply a softmax function to the representation of each node. That is, given a total of $\ell$ layers,
\begin{align} 
\textstyle\vec{z}_{v,k} = \textsc{Softmax} \left(\vec{h}_{v,k}^{\ell} \right)=
\frac{\exp \left(\vec{h}_{v,k}^{\ell} \right)}{\sum_{k'=1}^{K}\exp \left(\vec{h}_{v,k'}^{\ell} \right)}.
\end{align}

For a labeled node $v\in V_Y$, let $Y_{v,k} \in \{0,1\}$ be 1 if and only if  node $v$ belongs to class $k$. The overall loss is then formulated using a cross-entropy loss with regularization as
\begin{align}\label{eq:loss}
    -&\textstyle\sum_{v\in V_Y}\sum_{k=1}^K Y_{v,k} \ln \vec{z}_{v,k} + \lambda_G \|\Theta_{G}\|^2_2 +\textstyle\lambda_{L}\|\Theta_L\|_2^2\nonumber\\
  +& \lambda_{}\left(  {\|A - 1\|_2^2}/{|A|} + {\|B\|_2^2}/{|B|} \right),
\end{align}
where $\Theta_{G}=\{\vec{W}^{l}: l\le\ell\}$ contains the parameters of the global model, 
$\Theta_{L}=\{\vec{M}_a^{l},\vec{M}_b^{l}, \vec{N}_a^{l}, \vec{N}_b^{l}: l\le\ell\}$ contains the parameters of localization,
and $A$, $B$ are sets respectively containing the transformation vectors for scaling ($\vec{a}^{l}_{v}$'s and $\vec{a}^{l}_{  u,v }$'s) and shifting ($\vec{b}^{l}_{v}$'s and $\vec{b}^{l}_{  u,v }$'s). 
Note that the norms $\|\cdot\|^2_2$ are computed over all tensor elements in the sets.
While $A$ and $B$ are not learnable, they are functions of the learnable $\Theta_L$. We explicitly constrain them to favor small local changes, \ie, close-to-one scaling and close-to-zero shifting. Moreover, as $A$ and $B$ grow with the size of the graph, their norms are further normalized by the total number of tensor elements in them, denoted by $|A|$ and $|B|$, respectively.
$\lambda_G$, $\lambda_L$ and $\lambda$ are non-negative hyperparameters.

\subsection{Connections to Other Works}\label{sec:model:connections}

Next, we discuss how LGNN is related to other lines of work in GNNs, network embedding and hypernetworks.

\paragraph{Relationship with existing GNNs.}
The proposed LGNN is able to localize any GNN model that follows the scheme of recursive neighborhood aggregation in Eq.~\eqref{eq:gnn}.
In particular, LGNN is a generalized GNN that subsumes, as special limiting cases, several conventional GNNs by adopting an appropriate aggregation function  and regularization.

Specifically, as $\lambda\to \infty$ in Eq.~\eqref{eq:loss}, the scaling and shifting would approach 1's and 0's in the limiting case, respectively. This is equivalent to removing all node-wise localization at both node and edge levels, where all nodes assume a global model. Thus, 
LGNN would be asymptotically equivalent to GCN \cite{kipf2016semi} and GIN \cite{xu2018powerful}, when using a graph convolution aggregator (roughly equivalent to the mean pooling) and an MLP aggregator on the sum-pooled representations, respectively.
Furthermore, with an appropriate regularization, LGNN can be asymptotically reduced to GAT\footnote{Here we only discuss GAT with one attention head.} \cite{velivckovic2017graph}. Consider the vectors for scaling in $A$, which can be split into the node-specific $A_V$ (containing $\vec{a}^{l}_{v}$'s) and edge-specific $A_E$ (containing $\vec{a}^{l}_{  u,v }$'s). 
For $A_V$, we maintain the same regularization $\|A_V-1\|^2_2$; for $A_E$, we adopt the regularization $\|A_E-\alpha\|^2_2$ such that $\alpha$ contains the corresponding attention coefficients on all edges.
That is, 
\begin{align}
    \vec{a}^{l}_{  u,v  }\to \left[\left(\alpha^{l}_{  u,v  }\right)_{\times d_{l}}\right]^\top,
\end{align}
where $\alpha^{l}_{  u,v  } \in \mathbb{R}$ is the attention coefficient on edge $\langle u,v\rangle$.
Coupled with a mean aggregator, we obtain GAT as the limiting case when $\lambda\to \infty$.
Alternatively, we can also maintain the same regularization for both node- and edge-specific transformations, and adopt a  weighted mean aggregator in accordance to  the attention coefficients.
Note that in GAT the attention coefficient  
on each edge is a scalar, whereas in our edge-level localization, the edge-specific scaling is represented by a vector, which is able to flexibly vary the contribution from different dimensions of the information received through the edges. 

\paragraph{Relationship with network embedding.}
Network embedding \cite{cai2018comprehensive} is a popular alternative to GNNs to learn node representations on a graph. In these methods, each node is directly encoded with a learnable parameter vector, which is taken as the representation of the node. In contrast, in GNNs, the representations of the nodes are derived from a shared learnable model. Thus, network embedding  can be viewed as a set of individual local models that are loosely coupled by local graph structures, whereas GNNs capture a more global view of the structures. In contrast, the proposed LGNN achieves a balance between the local and global views, to allow localized variations grounded on a global model. 

\paragraph{Relationship with hypernetworks.}
Our localization strategy can be deemed a form of hypernetworks \cite{ha2016hypernetworks},
which use a secondary neural network to generate weights for the target network. In our case, we employ dense layers as a secondary network to generate the vectors for scaling and shifting, which are leveraged to ultimately generate the localized weights for the target GNN. 
Moreover, our approach boils down to the feature-wise modulation of information received from the conditioning local context (\ie, neighboring nodes) at both node and edge levels, 
which is inspired by the feature-wise modulation of neural activations in FiLMs \cite{perez2018film,birnbaum2019temporal}.

On graphs, GNN-FiLM \cite{brockschmidt2019gnn} also uses a form of FiLM on GNNs. However, there are several fundamental differences between GNN-FiLM and our LGNN. 
First, in terms of \emph{problem and motivation}, 
GNN-FiLM aims to model edge labels on relational networks for message passing. In contrast, our LGNN aims to model local contexts for node-wise localization, and works on general graphs. 
Second, in terms of \emph{model}, GNN-FiLM models
relation-specific transformations conditioned only on a node's self-information, which does not sufficiently reflect the full local context of the node as LGNN is conditioned on.
Furthermore, when there is no edge labels, GNN-FiLM reduces to a ``uniform edge" model. In LGNN, we still have both node and edge level modulation to achieve localization. 
Third, in terms of \emph{empirical performance}, 
as discussed therein \cite{brockschmidt2019gnn}, GNN-FiLM only achieves at best comparable performance to existing GNNs on citation networks (such as \emph{Cora} and \emph{Citeseer}) where there is no edge label. 
Our own experiments also have reproduced similar results in the next section.

\section{Experiments}

In this section, we evaluate and analyze the empirical performance of our proposed approach LGNN.


\subsection{Experimental Setup\protect\footnote{Additional implementation details and experimental settings are included in Sections A and B of the supplemental material. 
}}


\paragraph{Datasets.}
We utilize four benchmark datasets. They include two academic citation networks, namely \emph{Cora} and \emph{Citeseer} \cite{yang2016revisiting}, in which nodes correspond to papers and edges correspond to citations between papers. The input node features are bag-of-word vectors, indicating the presence of each keyword. A similar citation  network for Wikipedia articles called \emph{Chameleon} \cite{Rozemberczkimusae} is also used.
Finally, we use an e-commerce co-purchasing network called \emph{Amazon} \cite{hou2020measuring}, in which the nodes correspond to computer products and the edges correspond to co-purchasing relationships between products. The input node feature vectors are constructed from product images.
The statistics of the datasets are summarized in Table~\ref{table:datasets}.

\begin{table}[!t]
\centering
\footnotesize
\addtolength{\tabcolsep}{-0.5pt}
\begin{tabular}{@{}c|rrrrr@{}}
 \toprule
 Dataset	&   \# Nodes 	& \# Edges 	&  \# Classes  &  \# Features  \\
\midrule
 Cora  & 2,708  &  5,429  &  7  &  1,433 \\
 Citeseer & 3,327  &  4,732  &  6  &  3,703 \\ 
 Amazon  & 13,381  &  245,778  &  10  &  767 \\
 Chameleon  & 2,277  &  36,101  &  5  &  2,325 \\\bottomrule 
\end{tabular}
\caption{Summary of datasets.\label{table:datasets}}
\end{table}

\begin{table*}[!t] 
    \centering
    \footnotesize
    \begin{tabular}{@{}l|r|cc|cc|cc|cc@{}}
    \toprule
   \multirow{2}*{Methods} &
   {\# Params} & \multicolumn{2}{c|}{Cora} & \multicolumn{2}{c|}{Citeseer} &  \multicolumn{2}{c|}{Amazon} & \multicolumn{2}{c}{Chameleon} \\
      & (Cora) & Accuracy & Micro-F  & Accuracy & Micro-F & Accuracy & Micro-F & Accuracy & Micro-F \\\midrule \midrule
     DeepWalk & 693K & 73.8$\pm$0.3 & 74.9$\pm$0.1 & 61.6$\pm$0.2 & 60.5$\pm$1.0 &   80.1$\pm$1.6 & 77.3$\pm$1.3 & 41.2$\pm$1.3 & 40.1$\pm$1.1 \\
     Planetoid & 345K& 66.1$\pm$0.4 & 64.5$\pm$0.5 & 64.5$\pm$0.3 & 62.9$\pm$0.4 & 69.8$\pm$1.7 & 64.5$\pm$1.5 & 39.3$\pm$1.8 & 37.7$\pm$1.7  \\ \midrule
     GCN& 11K & 81.5$\pm$0.7 & 80.8$\pm$0.5  & 70.4$\pm$0.5 & 68.3$\pm$0.7  &  81.9$\pm$0.5 & 81.0$\pm$0.8 & 46.7$\pm$4.3 & 46.4$\pm$2.4    \\
     GCN-64 & 92K & 82.0$\pm$0.3 & 80.9$\pm$0.3  & 71.1$\pm$0.3 & 69.2$\pm$0.4  &  82.1$\pm$0.5 & 81.2$\pm$0.8 & 48.3$\pm$3.3 & 46.3$\pm$1.8  \\
     GCN-96& 138K & 81.9$\pm$0.2 & 80.8$\pm$0.3  & 71.3$\pm$0.4 & 69.4$\pm$0.5  &  82.2$\pm$0.4 & 81.5$\pm$0.7 & 45.5$\pm$2.4 & 43.8$\pm$2.5   \\
     GCN-FiLM& 35K & 78.1$\pm$0.6 &  76.9$\pm$0.5 & 69.8$\pm$1.1  & 67.9$\pm$1.0 & 79.2$\pm$1.0  & 77.1$\pm$1.5 & 42.8$\pm$1.1 & 39.9$\pm$1.3   \\
     LGCN& 104K & \textbf{83.5}$\pm$0.3 & \textbf{82.1}$\pm$0.4 & \textbf{72.2}$\pm$0.4 & \textbf{70.2}$\pm$0.4 &  \textbf{83.7}$\pm$1.5 & \textbf{82.3}$\pm$2.0 & \textbf{50.9}$\pm$1.1 & \textbf{49.7}$\pm$0.7   \\
     (improv.) & - &(1.8\%) & (1.5\%) &	(1.3\%) & (1.2\%) &  (1.8\%) & (1.0\%) &  (5.4\%) & (7.1\%) \\
     \midrule
     GAT& 92K & 82.9$\pm$0.6 & 82.0$\pm$0.6 &  72.4$\pm$0.7 & 70.4$\pm$0.8 &  82.4$\pm$1.3 & 80.1$\pm$1.9 &  47.2$\pm$1.1 & 46.2$\pm$2.1   \\
     GAT-64& 738K & 83.1$\pm$0.4 & 81.9$\pm$0.6 &  71.6$\pm$1.5 & 69.8$\pm$1.6 &  83.0$\pm$0.9 & 81.2$\pm$1.4 &  51.2$\pm$1.5 & 50.2$\pm$1.3   \\
     GAT-96& 1108K & 83.2$\pm$0.6 & 81.9$\pm$0.6 &  71.4$\pm$0.9 & 69.6$\pm$0.9 &  83.1$\pm$1.0 & 81.5$\pm$1.4 &  51.9$\pm$1.2 & 50.2$\pm$1.8   \\
     GAT-FiLM& 277K & 82.0$\pm$0.5 & 80.6$\pm$0.6  & 71.2$\pm$1.0 & 69.2$\pm$1.1  &  83.3$\pm$0.6 & 81.9$\pm$0.8 & 46.8$\pm$5.7 & 45.1$\pm$5.2   \\
    LGAT& 836K & \textbf{83.6}$\pm$0.4 & \textbf{82.3}$\pm$0.4 & \textbf{72.8}$\pm$0.4 & \textbf{70.8}$\pm$0.5 &  \textbf{83.7}$\pm$0.7 & \textbf{82.3}$\pm$0.8 &  \textbf{52.6}$\pm$1.0 & \textbf{51.1}$\pm$0.9  \\
     (improv.) & - &(0.5\%) & (0.4\%) &	(0.6\%) & (0.6\%) &  (0.5\%) & (0.5\%) &  (1.3\%) & (1.8\%) \\
     \midrule
    GIN& 11K & 80.2$\pm$0.5 & 78.8$\pm$0.3 &  68.5$\pm$0.7 & 66.5$\pm$1.0 &  79.6$\pm$1.7 & 78.5$\pm$2.6 &  45.8$\pm$3.0 & 41.2$\pm$4.0   \\
     GIN-64& 92K & 80.3$\pm$1.1 & 79.1$\pm$1.0 &  67.8$\pm$1.5 & 66.1$\pm$1.1 &  79.8$\pm$1.1 & 79.0$\pm$1.4 &  45.7$\pm$4.5 & 40.7$\pm$5.7   \\
     GIN-96& 138K & 79.9$\pm$1.1 & 78.9$\pm$1.0 &  68.6$\pm$1.4 & 66.6$\pm$1.6 &  80.2$\pm$2.1 & 79.0$\pm$3.2 &  45.9$\pm$3.5 & 41.5$\pm$4.1   \\
     GIN-FiLM& 35K & 79.8$\pm$0.7 & 78.5$\pm$0.5  & 67.7$\pm$1.4 & 65.8$\pm$1.5  &  78.6$\pm$2.8 & 77.2$\pm$3.3 & 38.8$\pm$2.6 & 34.2$\pm$2.9   \\
    LGIN& 126K & \textbf{82.6}$\pm$0.8 & \textbf{81.6}$\pm$0.8 &  \textbf{71.3}$\pm$0.4 & \textbf{69.5}$\pm$0.5 &  \textbf{84.0}$\pm$1.2 & \textbf{82.7}$\pm$1.7 & \textbf{48.3}$\pm$1.9 & \textbf{47.3}$\pm$1.9  \\
     (improv.) & - &(2.9\%) & (3.2\%) &	(3.9\%) & (4.4\%) &  (4.7\%) & (4.7\%) &  (5.2\%) & (14.0\%) \\ 
     \bottomrule
     \end{tabular}
     \caption{Average classification performance with standard deviation (percent) over 10 runs. 
    Improvements of LGNN are  relative to the best performing baseline with the corresponding GNN architecture.
    }
    \label{table:results-20}%
\end{table*}

\paragraph{Baselines and our approach.}
We compare against three categories of baselines. (1) Embedding models: \emph{DeepWalk} \cite{perozzi2014deepwalk} and \emph{Planetoid} \cite{yang2016revisiting}. Both employ a direct embedding lookup and adopt random walks to sample node pairs. However, 
DeepWalk is unsupervised such that node classification is performed as a downstream task on the learned representations, whereas
Planetoid is an end-to-end model that jointly learns the representations and the classifier.
(2) GNN models: \emph{GCN} \cite{kipf2016semi}, \emph{GAT} \cite{velivckovic2017graph} and \emph{GIN} \cite{xu2018powerful}. 
(3) GNN-FiLM: The original GNN-FiLM works with a GCN-style model, which we call \emph{GCN-FiLM}. We further extend it to two other GNN architectures GAT and GIN, resulting in two more versions \emph{GAT-FiLM} and \emph{GIN-FiLM}.

On the other hand, our approach LGNN can also be implemented on different GNN architectures. Specifically, we employ GCN, GAT and GIN as the global model, and obtain localized versions \emph{LGCN}, \emph{LGAT} and \emph{LGIN}, respectively.

\paragraph{Main settings.}
For DeepWalk, we sample 10 random walks per node with walk length 100 and windows size 5, and set the embedding dimension to 128. We further train a logistic regression as the downstream classifier. For Planetoid, we set the path length to 10, window size to 3 and the embedding dimension to 50. We select the best results from their transductive and inductive versions for report.
For all GNN-based approaches, we adopt two aggregation layers, noting that deeper layers often bring in noises from distant nodes \cite{pei2020geomgcn} and cause ``over-smoothing'' such that all nodes obtain very similar representations \cite{li2018deeper,pmlr-v80-xu18c}. The dimension of the hidden layer defaults to 8, while we also present results using larger dimensions. The regularization of GNN parameters is set to $\lambda_G=0.0005$. 
These settings are chosen via empirical validation, and are largely consistent with the literature \cite{perozzi2014deepwalk,kipf2016semi,velivckovic2017graph}.
For our models, the additional regularizations are set to $\lambda_L=\lambda=1$ (except for $\lambda=0.1$ in LGAT). 

\paragraph{Training and testing.}
For all datasets, we follow the standard split in the literature \cite{yang2016revisiting,kipf2016semi,velivckovic2017graph}, which uses 20 labeled nodes \emph{per class} for training, 500 nodes for validation and 1000 nodes for testing. 
We repeat 10 runs for each experiment, and report the average accuracy and micro-F score. 

\begin{figure*}[t]
\centering
\includegraphics[width=0.95\linewidth]{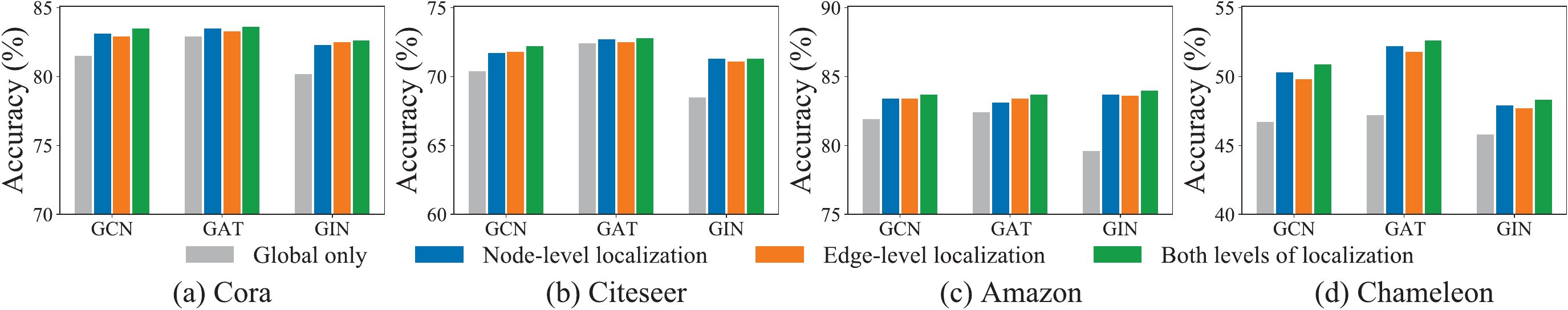}
\caption{Ablation study on the effect of each localization module.}
\label{fig.ablation}
\end{figure*}

\subsection{Performance Comparison}

Table~\ref{table:results-20} shows the comparison of the classification performance on the four datasets. 
For each GNN architecture, we also include additional models by increasing the dimension of its hidden layer. For instance, GCN-64 denotes that the hidden layer of GCN has 64 neurons. 
The results reveal a few important observations.

\emph{Firstly}, we observe that LGNN consistently achieves significant performance boosts \wrt three state-of-the-art GNNs on four datasets.
This suggests that, our localization approach is agnostic of the particular GNN architecture of the global model, and can be used universally on different GNNs. 
Furthermore, LGNNs also outperform GNN-FiLMs, as GNN-FiLMs only achieve at best comparable performance with the base GNNs. A potential reason is that GNN-FiLM is only conditioned on the target node's self-information to modulate message passing for different edge labels, which may not be useful on graphs without edge labels. In contrast, LGNN is conditioned on the full local context of the target node to achieve node-wise localization that could be useful to the classification of nodes residing across different localities on the graph.
\emph{Secondly}, GAT-based models generally attain better performance than GCN- and GIN-based models, and the performance gain in LGAT w.r.t.~GAT is smaller. This is not surprising as the attention coefficients on the edges can be understood as a limited form of localization to differentiate the weight of different neighbors.
\emph{Thirdly}, the results imply that increasing the number of parameters alone cannot achieve the effect of localization. It may be argued that more parameters lead to higher model capacity, which can potentially assign some space to encode the local aspects. However, when the parameters are still shared globally, there is no explicit node-wise constraint and thus the model will still try to find a middle ground. Nevertheless, for a fair comparison, we use more parameters on each GNN baseline to match or exceed the number of parameters on the corresponding LGNN, as listed in
Table~\ref{table:results-20}\footnote{Details of the calculation are included in Section C of the supplemental material.}. 
As expected, more parameters only result in marginal improvements to the baselines.
Thus, the efficacy of LGNN is derived from our localization strategy rather than just more parameters. 
\emph{Fourthly}, LGNN is robust and stable given their relatively small standard deviations in many cases. We hypothesize that, in conventional GNNs, a bad initialization would risk a suboptimal global model shared by all nodes without recourse. In contrast, node-wise localization offers an opportunity to adjust the suboptimal model for some nodes, and is thus less susceptible to the initialization.

It has also been discussed that in GNNs overfitting to the validation set is an issue \cite{shchur2018pitfalls}, and thus models with more parameters tend to perform  better given a larger validation set. Therefore, we further compare different methods using a smaller validation set of only 100 nodes\footnote{These results are  in Section D of
the supplemental material.}. Our proposed LGNN still outperforms all the baselines across the four datasets, demonstrating its power and stability.

\subsection{Further Model Analysis}

Due to space constraint, we only present an ablation study here\footnote{More analyses on the complexity and effect of regularization, as well as model visualization, can be found in Sections E, F and G of the supplemental material.}.
In the ablation study, we investigate the effectiveness of the node- and edge-level localization modules in LGNN.
To validate the contribution of each  module, we compare the accuracy of four variants in Fig.~\ref{fig.ablation}: (1)
global only without any localization; (2) node-level localization only; (3) edge-level localization only; and (4) the full model with both localization modules. 
We observe that utilizing only one module, whether at the node or edge level, consistently outperforms the global model. Between the two modules, the node-level localization tends to perform better.  
Nevertheless, modeling both jointly results in the best performance, which implies that both modules are effective and possess complementary localization power.   

\section{Related Work}

Graph representation learning has received significant attention in recent years. 
Earlier network embedding approaches \cite{perozzi2014deepwalk,tang2015line,grover2016node2vec} employ a direct embedding lookup for node representations, which are often coupled via local  structures such as skip-grams.
To better capture global graph structures, GNNs \cite{kipf2016semi,hamilton2017inductive,velivckovic2017graph,xu2018powerful} open up promising opportunities. They generally follow a paradigm of recursive neighborhood aggregation, in which each node receives information from its neighbors on the graph.
More complex structural information is also exploited by recent variants \cite{zhangkai2020adaptive,pei2020geomgcn,chen2020simple},
but none of them deals with the node-wise localization.

While our approach can be deemed a form of hypernetworks \cite{ha2016hypernetworks,perez2018film} as discussed in Sect.~\ref{sec:model:connections},
several different forms of local models have been explored in related problems. In manifold learning \cite{yu2009nonlinear,ladicky2011locally}, local codings are used to approximate any point on the manifold as a linear combination of its surrounding anchor points. In low-rank matrix approximation \cite{lee2013local}, multiple low-rank approximations  are constructed for different regions of the matrix before 
aggregation.
These methods are tightly coupled with their problems, and cannot be easily extended to localize GNNs.
Some meta-learning frameworks \cite{vinyals2016matching,snell2017prototypical,finn2017model} can also be viewed as adapting to local contexts. Given a set of training tasks, meta-learning aims to learn a prior that can be adapted to new unseen tasks, often for addressing few-shot learning problems.  
In particular, existing meta-learning models on graphs are mostly designed for few-shot node classification \cite{zhou2019meta,yao2019graph,liu2021relative}
or regression \cite{liu2020towards}. 
Such few-shot settings are fundamentally different from our localization objective, in which all nodes are seen during training and we leverage the local contexts of the nodes in order to enhance their representations. 

\section{Conclusions}

In this work, we identified the need to localize GNNs for different nodes that reside in non-uniform local contexts across the graph. 
This motivated us to propose a node-wise localization approach, named LGNN, in order to adapt to the locality of each node.
On one hand, we encode graph-level general patterns using a global weight matrix.
On the other hand, we modulate the global model to generate localized weights specific to each node,
and further perform an edge-level modulation to enable finer-grained localization.
Thus, the proposed LGNN can capture both local and global aspects of the graph well. 
Finally, our extensive experiments demonstrate that LGNN significantly outperforms state-of-the-art GNNs.

\section*{Acknowledgments}
This research is supported by the Agency for Science, Technology and Research (A*STAR) under its AME Programmatic Funds (Grant No. A20H6b0151).

\newpage


\bibliographystyle{named}

\end{document}